%% file: aaai25.tex
\documentclass[letterpaper]{article} 
\usepackage{aaai25}  
\usepackage{times}  
\usepackage{helvet}  
\usepackage{courier}  
\usepackage[hyphens]{url}  
\usepackage{graphicx} 
\usepackage{natbib}  
\usepackage{caption} 
\usepackage{algorithm}
\usepackage{algorithmic}
\usepackage{url}

\usepackage{newfloat}
\usepackage{listings}
\DeclareCaptionStyle{ruled}{labelfont=normalfont,labelsep=colon,strut=off} 
\floatstyle{ruled}
\newfloat{listing}{tb}{lst}{}
\floatname{listing}{Listing}
\title{“From Unseen Needs to Classroom Solutions”: Exploring AI Literacy Challenges \& Opportunities with Project-Based Learning Toolkit in K-12 Education}
\author {
    Hanqi Li\equalcontrib\textsuperscript{\rm 1},
    Ruiwei Xiao\equalcontrib\textsuperscript{\rm 2},
    Hsuan Nieu\textsuperscript{\rm 3},
    Ying-Jui Tseng\textsuperscript{\rm 2},
    Guanze Liao\textsuperscript{\rm 3}
}
\affiliations {
    \textsuperscript{\rm 1}New York University\\
    \textsuperscript{\rm 2}Carnegie Mellon University\\
    \textsuperscript{\rm 3}Taiwan National Tsing Hua University\\
    hl4893@nyu.edu, ruiweix@andrew.cmu.edu, dotnew@gapp.nthu.edu.tw, yingjuit@andrew.cmu.edu, gzliao@mx.nthu.edu.tw
}

\begin{document}

\maketitle

\input{Sections/00_Abstract}
\input{Sections/01_Introduction}
\input{Sections/02_Related_Works}
\input{Sections/03_Toolkit}

\input{Sections/04_Method}

\input{Sections/05_Results}
\input{Sections/06_Discussion}

\bibliography{aaai25}

\end{document}

%% file: Sections/00_Abstract.tex
\begin{figure*}[ht]
  \centering
  \includegraphics[width=1\textwidth]{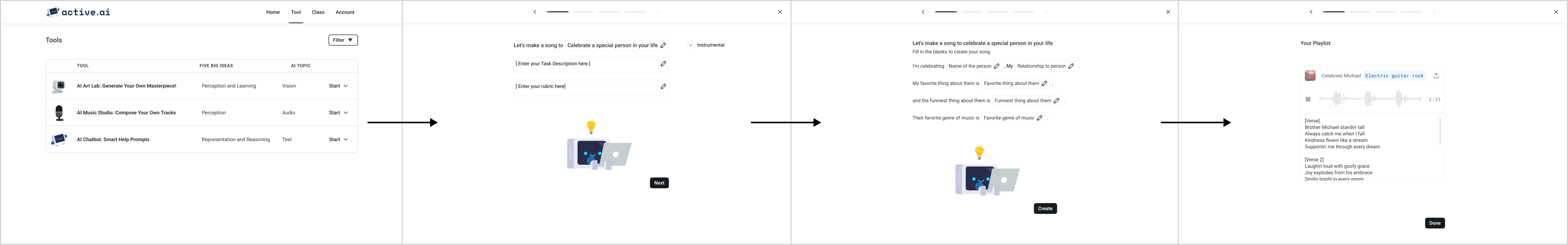} 
  \caption{The AI Music Studio, one of our three Project-Based Learning (PBL) tools, enables students to generate customized music based on teacher-provided rubrics. (From left to right) Students first define the task objective and input the rubric, then personalize the content with key details (e.g., name, relationship, favorite qualities, and music genre). The system generates a song with audio playback and lyrics reflecting their inputs. A demo can be accessed on Figma (\url{https://shorturl.at/Ra5EQ}).
}
  \label{fig:ux_design}
\end{figure*}

\begin{abstract}
As artificial intelligence (AI) becomes increasingly central to various fields, there is a growing need to equip K-12 students with AI literacy skills that extend beyond computer science. This paper explores the integration of a Project-Based Learning (PBL) AI toolkit into diverse subject areas, aimed at helping educators teach AI concepts more effectively. Through interviews and co-design sessions with K-12 teachers, we examined their current AI literacy levels and how these teachers adapt AI tools like the AI Art Lab, AI Music Studio, and AI Chatbot into their course designs. While teachers appreciated the potential of AI tools to foster creativity and critical thinking, they also expressed concerns about the accuracy, trustworthiness, and ethical implications of AI-generated content. Our findings reveal the challenges teachers face, including limited resources, varying student and instructor skill levels, and the need for scalable, adaptable AI tools. This research contributes insights that can inform the development of AI curricula tailored to diverse educational contexts. 

\end{abstract}

%% file: Sections/01_Introduction.tex
\section{Introduction}
As accessible Artificial Intelligence (AI) tools have gained increasing interest among K-12 educators in incorporating AI literacy into their classrooms. K-12 educators recognize the need to teach students about its capabilities and limitations\cite{ng2023ai}. Existing AI education efforts focus on dedicated curricula and professional learning for teachers \cite{amplo2023learning, lee2022preparing}. However, many teachers face challenges, such as limited AI knowledge, time constraints, and resource availability, especially outside of computer science classes \cite{walsh2023literacy,song2022paving}. Scalable and adaptable resources are needed to help teachers incorporate AI into their lessons without requiring deep technical expertise. To address these challenges, the Active AI team explored the development of modular AI resources, combined with Project Based Learning (PBL), which has brought positive feedback in student learning \cite{kong2024developing}, so that teachers can adapt to their subjects, allowing them to introduce AI literacy in more flexible units \cite{tseng2024activeai}. We conducted interview and co-design sessions with teachers from various backgrounds to investigate three key questions: 

\begin{itemize}

\item \textbf{RQ1:} What is the AI literacy level of K-12 teachers?

\item \textbf{RQ2: }How do IT and interdisciplinary course instructors design lesson plans using our PBL toolkit?

\item \textbf{RQ3:} How do differences in teachers’ and students’ backgrounds influence their perspectives on AI tools, student performance, and course design? 
\end{itemize}
 
To answer these questions, this paper presents a formative empirical evaluation of teachers' perspectives on AI literacy, providing insights that can inform the development of adaptable AI tools to support AI literacy education across diverse subject areas.

%% file: Sections/02_Related_Works.tex
\section{Related Works}
\subsection{AI Literacy Curriculum}

As artificial intelligence (AI) becomes increasingly integral to a wide array of applications, it is essential for K–12 students to understand how to interact with these technologies and recognize their capabilities and limitations \cite{ng2023ai, eguchi2021contextualizing}. Prior work has investigated AI curricula beyond computer science and programming, focusing on creative, human-centered aspects of AI \cite{eaton2018blue, walsh2023literacy}. These cross-disciplinary AI curricula aim to make AI education more accessible to students from various backgrounds.

Many researchers have launched co-design collaborations to develop AI literacy curricula \cite{williams2023ai+, amplo2023learning, bilstrup2022supporting}.  \citeauthor{amplo2023learning} highlighted the importance of profound collaborations with teachers, where educators play an active role in shaping the curriculum while gaining more knowledge about the field. Such partnerships empower teachers as creators that could tailor new curricula to student needs and interests, instead of just implementers of AI curricula \cite{lee2022preparing, dipaola2023make}. 

For example, \citeauthor{walsh2023literacy} found that literacy and STEM teachers with minimal AI experience were able to adapt an AI ethics curriculum, suggesting that AI knowledge is not a prerequisite for curricular co-design.  Co-designing AI education curricula with cross-disciplinary high school teachers has been shown to enhance the relevance and adaptability of the curriculum \cite{lin2021engaging}. \citeauthor{lin2021engaging} also engaged teachers to co-design integrated AI curricula for K-12 classrooms and found that teachers need additional scaffolding in AI tools and curriculum to facilitate ethics and data discussions, and value supports for learner evaluation and engagement, peer-to-peer collaboration, and critical reflection.

\subsection{Project-Based Learning in AI Literacy}

Project-Based Learning (PBL) is an active, student-centered form of instruction characterized by autonomy, constructive investigations, goal-setting, collaboration, communication, and reflection within real-world practices \cite{kokotsaki2016project, condliffe2017project}. Collaborative PBL, which combines interdisciplinary problem-solving and artifact creation, is a common approach in AI literacy education. \cite{ng2023artificial,tseng2024assessing}.

Several studies have implemented PBL in AI literacy education with positive outcomes. \citeauthor{kong2024developing} developed an AI literacy framework and evaluated a literacy course for senior secondary students using a PBL approach, finding that students improved in AI problem-solving ability, empowerment, and ethical understanding. \citeauthor{jeon2023developing} investigated middle school students' AI literacy through a PBL program and confirmed efficacy in both cognitive and affective gains. \citeauthor{ng2024fostering} fostered secondary school students’ AI literacy through making AI-driven recycling bins in a PBL intervention, demonstrating positive impacts on motivation, AI literacy, and collaboration.

However, many AI literacy PBL implementations are not scalable due to resource constraints, lack of trained educators, and the complexity of AI topics \cite{walsh2023literacy}. \citeauthor{song2022paving} highlighted that AI knowledge and technical skills are limited not only for students but also for school teachers, especially in rural areas. Druga et al. \cite{druga2022landscape} proposed that future AI curricula move from singular activities and demos to more holistic designs that include support, guidance, and flexibility for how AI technology, concepts, and pedagogy play out in the classroom. Researchers also emphasized the need for sustainable approaches to inform the planning of quality AI curricula, as educators and experts have found designing AI-related curricula a challenge \cite{chiu2021holistic, cu2022ai}.

Our research aims to bridge this gap by developing a high-fidelity prototype that serves as a tangible interface for facilitating teacher interviews. The toolkit is designed to generate hands-on PBL AI tool exploration activities that teachers can seamlessly integrate into their lesson plans, regardless of their prior AI experience. This approach promotes accessible and sustainable AI literacy across diverse educational settings. It is important to note that while the toolkit focuses on hands-on AI tool exploration, corresponding knowledge-focused instruction is covered in other course modules within our overarching project \cite{xiao2024activeai}.

%% file: Sections/03_Toolkit.tex
\section{Toolkit}

\subsection{System Overview}
The Active AI team has integrated three AI demo tools into our PBL toolkit: AI Art Lab, AI Music Studio, and AI Chatbot (Figure \ref{fig:ux_design}). Each tool offers a complete user flow that our users can experiment with.
AI Art Lab utilizes an explained version of the image generation diffusion model, where teachers can set a topic or scope. Students can upload their own image datasets and, following step-by-step instructions, generate their own images.
AI Music Studio allows students to create music based on a teacher-assigned topic. With AI guidance, they can fill in details such as the theme and music style to generate their unique compositions.
AI Chatbot is an AI-driven conversation tool that allows students to interact with AI on predefined subjects and difficulty levels. Additionally, the AI can evaluate the quality of the prompts or questions students provide, helping them refine their inquiry skills.
These tools are designed to enhance both creativity and critical thinking, offering opportunities for both students and teachers to engage with a variety of interactive AI-based learning experiences.

\subsection{Design Rationale}

\subsubsection{PBL tools is good for instruction in AI Literacy Education}

We developed the PBL toolkit for AI literacy education to address the growing need for students to engage with AI across various disciplines and recognize its capabilities and limitations. As AI becomes integral to numerous fields, prior research has highlighted the importance of accessible, human-centered AI curricula that go beyond technical knowledge and reach students from diverse backgrounds. Collaborative approaches, particularly co-designing with teachers, have proven successful in tailoring AI education to meet student needs(Kokotsaki,
Menzies, and Wiggins 2016; Condliffe 2017). However, challenges such as resource limitations, the complexity of AI topics, and the varying levels of AI literacy among teachers, especially in underserved areas, make it difficult to implement AI curricula widely. By building a PBL toolkit, we aim to narrow these gaps, offering educators practical tools to integrate AI into their lesson plans, regardless of their AI expertise. The toolkit fosters hands-on, interdisciplinary learning, making AI literacy more accessible and sustainable across diverse educational contexts.

\subsubsection{Mapping to AK4K12 curriculum}
To align our AI tools in the toolkit (AI Art Lab, AI Music Studio, AI Chatbot) to the goals outlined in the AI4K12 curriculum, which focuses on Perception, Representation \& Reasoning, Learning, Natural Interaction, and Societal Impact. 

The AI tools in our toolkit align with the AI4K12 framework in several key areas. The AI Art Lab addresses Perception by allowing students to explore how AI perceives and generates images, while also promoting Learning as they engage with AI models to understand pattern recognition in art. The AI Music Studio similarly supports Perception, demonstrating how AI processes sound, and fosters Learning by training AI to create music, enhancing students' creativity. The AI Chatbot focuses on Representation \& Reasoning, helping students understand how AI processes language, and Natural Interaction, by enabling meaningful conversations with AI. Additionally, it encourages reflection on Societal Impact, prompting students to consider AI's role in communication and its broader implications.

%% file: Sections/04_Method.tex
\renewcommand{\arraystretch}{1.2} 
\begin{table*}[ht] 
\scriptsize 
\centering
\begin{tabular}{|l|p{1.2cm}|p{1cm}|p{1cm}|p{1cm}|p{10cm}|}
  \hline
  \textbf{ID} & \textbf{Subjects Taught} & \textbf{School Type} & \textbf{Years of Teaching} & \textbf{Location} & \textbf{Tool-Based Course Design} \\ 
  \hline
  P-GS-Z & General subject & Public & 2 & North America & \textbf{AI chatbot} to reimagine a romance story with \textit{Romeo and Juliet} themes, guided by editable task descriptions and customizable grading rubrics.\\ 
  \hline
  V-MA-Z & Media Art & Private & 1 & North America & Students use \textbf{AI chatbot} for AI Q\&A and interactive learning, which trains students to ask better questions and provides tailored feedback, versatile for subjects like language and science. \\ 
  \hline
  C-MathI-Z & Math, IB curriculum & Public (charter) & 5 & North America & Students use \textbf{AI chatbot} for question-asking and algorithmic thinking skills in mathematics and Theory of Knowledge, helping students refine inquiry and problem-solving techniques. \\ \hline
  P-GSci-X & General science & Public & 11 & North America & Students use \textbf{AI chatbot} for conversational engagement and concept reinforcement, offering a platform to discuss physics concepts, review notes, and clarify ideas in a supportive, non-judgmental environment. \\ 
  \hline
  P-CH-Z & Chinese & Public & 6 & North America & Students use the \textbf{AI Music Studio} tool to create songs incorporating newly learned Chinese vocabulary for family members, promoting active participation through music-based group activities and addressing previous learning challenges. \\ 
  \hline
  P-SS-Z & Social studies & Public & 2 & North America & Students use \textbf{AI chatbot} for feedback provision and creative project generation, while the \textbf{AI Art Lab} supports creative projects like generating posters and logos for a mock presidential campaign, introducing tools for art and design exploration.\\ 
  \hline
  P-GSci-Y & Algebra \& geometry & Public & 4 & North America & Students use the \textbf{AI Music Studio} to create songs integrating geometry concepts, differentiating shapes like squares and rectangles. Rubrics ensure the inclusion of shape types, properties, and a personal understanding of their favorite shape. \\ 
  \hline
  V-IT-Z & IT & Private & 3 & East Asia & Students use \textbf{AI chatbot} for their free exploration, with younger students engaging in personified interactions and older students receiving homework assistance. The tool supports ethical discussions and practical projects like board game design and video editing, effectively addressing previous challenges. \\ 
  \hline
  P-IT-Z & IT & Public & 18 & East Asia & Students use the \textbf{AI Art Lab} to generate images for design projects, integrating textbook content into creating visuals for products to sell in an app development project. \\ 
  \hline
  P-GS-XY & General subject & Public & 27 & East Asia & Students use \textbf{AI chatbot} for knowledge sharing and discussion facilitation in music (string instruments) and art. It supports discussions, instructions, and even art generation. \\ 
  \hline
  V-IT-Z-2 & IT & Private & 7 & East Asia & Students use \textbf{AI chatbot} for problem-solving and guided inquiry across various subjects. Adaptable for different age groups, it enables teachers to pose problems while students research and explore real-world solutions. \\ 
  \hline
  V-Bio-Z & Biology, STEM club & Private & 27 & East Asia & Students use the \textbf{AI Music Studio} for music-based memory aids, converting key concepts like the periodic table into songs to enhance memory retention and engage through sensory stimulation across subjects like Chemistry. \\ 
  \hline
  V-EngIT-Z & English, IT & Private & 6 & East Asia & Students use the \textbf{AI Art Lab} to break down the image generation process, learning to input data and generate big data insights. \\ 
  \hline
\end{tabular}
\caption{Description of Teacher Participants' Teaching Subjects, Schools, Years of Teaching, Locations and the Course They Designed Using Our Toolkit. Participant IDs follow this structure: {School Type: P for public, V for private, C for charter}-{Main Subject Taught}-{Title Status: X for schools with majority of students come from low-income families, Y for schools with a majority of immigrant students, Z for schools that do not fall under either category.}}
\label{tab:teacher-participants}
\end{table*}

\section{Study Design}
\subsection{Participants}
The study is designed to capture both qualitative and quantitative data to provide a comprehensive understanding of the toolkit's implementation in classrooms and to assess participants' AI literacy. We recruited 13 middle and high school teachers who taught mixed subjects. Of these, 54\% were teaching in schools in North America and 46\% in East Asia, with 62\% teaching in public schools and 38\% in private schools. The participants had an average of 9.15 years (STDV=9.10) of teaching experience in 9 different subjects. 30\% of the teachers had experience teaching Information Technology (IT) courses, a prevalent subject in East Asian K-12 education systems that emphasizes computer usage, programming, and AI literacy, representing a diverse cross-section of educators with varying levels of experience in AI technologies and pedagogy. Participants were selected through purpose-sampling to ensure diverse teaching backgrounds and contexts. Before starting the study, participants were informed about the research goals, procedures, and data privacy protocols. Institutional Review Board (IRB) approval was obtained prior to data collection, and each participant provided informed consent. Demographic data was collected prior to the demo and interviews, which were conducted in English and Mandarin. Teachers received \$20 for participating in a one-hour interview, and all 13 sessions were analyzed.

\subsection{Procedure}
Participants started with watching a 5-minute video that demonstrated the PBL AI Literacy toolkit, highlighting its key features. The toolkit includes interactive AI tools across multiple modalities, such as a Large Language Model (LLM) chatbot, music generation, and image generation, allowing teachers to customize projects, scopes, and rubrics. After the video, participants explored the demo and then participated in a 50-minute interview. Interview questions covered four areas: 1) Level of AI literacy, 2) AI Literacy teaching experiences, 3)  Features in the toolkit, and 4) Course design with the toolkit. These questions prompted reflection on the educational challenges, the adaptability of the toolkit, and classroom implementation. Interviews were conducted via Zoom, connecting teachers from various regions with the research team. Each interview was facilitated by at least one researcher, and responses were recorded, transcribed, and analyzed using Zoom scripts and Google Sheets to identify themes related to AI tool application, adoption barriers, and opportunities for curriculum innovation.

%% file: Sections/05_Results.tex
\section{Results}

In this section, two researchers analyzed 13 interview scripts, totaling 14 hours of Zoom recordings to answer our research questions. Detailed demographic information and teachers' course designs can be found in Table \ref{tab:teacher-participants}.

\subsection{RQ1: What is the current level of AI literacy of K-12 teachers?} Among the teachers we interviewed, 38\% of teachers stated that they have a strong understanding of the AI4K12 framework or the 5 Big Ideas \cite{touretzky2023machine}, or have heard of similar frameworks \cite{long2020ai}; \textbf{69.23\% are confident} (self-rated confidence level $ \geq $ 50\%) \textbf{ in understanding the results of AI and recognizing their limitations}. Three teachers specifically mention that they are confident in identifying AI-generated student work due to their teaching experience, while others attribute their confidence to having used AI tools before. These teachers also recognize the fast-paced AI technology and the possibility of gaps in their knowledge.

\begin{quote}
    \scriptsize\textit{C-MathI-Z: “I feel fairly confident that I knew what was going on 6 months ago, but not so confident right now, and no idea what's coming in the next year... I feel pretty good, I feel pretty confident, but I know that I might be totally wrong.”}
\end{quote}

They further discussed their understanding of AI’s capabilities and limitations in education. The most commonly mentioned AI capability is \textbf{its potential to serve as a self-studying assistant to students}, whether as a conversational tutoring tool \cite{stamper2024enhancing} or by providing drafts and initial prompts for student projects. On the other hand, teachers noted that \textbf{AI’s results are not fully accurate}, and \textbf{ cannot replace students' critical and creative thinking}. Additionally, AI cannot replace the teacher’s role in real-life settings, such as being a role model and managing the classroom. 
\textbf{Over 92\% of teachers believe that students should use AI to support learning.} Most expressed a strongly positive attitude towards AI, citing reasons such as AI tools that enhance self-study when teachers cannot attend to all students, the widespread adoption of AI, and the importance of students keeping up with current trends.

\begin{quote}
    \scriptsize\textit{P-GSci-X : “I think that they should. Yes... it's a valuable skill to learn how to use AI. So every time that they use it, they're not just learning about the content that they're trying to engage, but they're also learning about how to engage with AIs.”}
\end{quote}

Furthermore, many teachers emphasized that AI education is essential to teach students how to \textbf{use these tools properly}. Some teachers, while generally positive, expressed concerns about the potential for misuse. 

\subsection{RQ2: How do IT and interdisciplinary course instructors design lesson plans using our PBL toolkit? }

\subsubsection{2.1 What course activities teachers designed?}

\begin{figure*}[ht]
  \centering
  \includegraphics[width=1\textwidth]{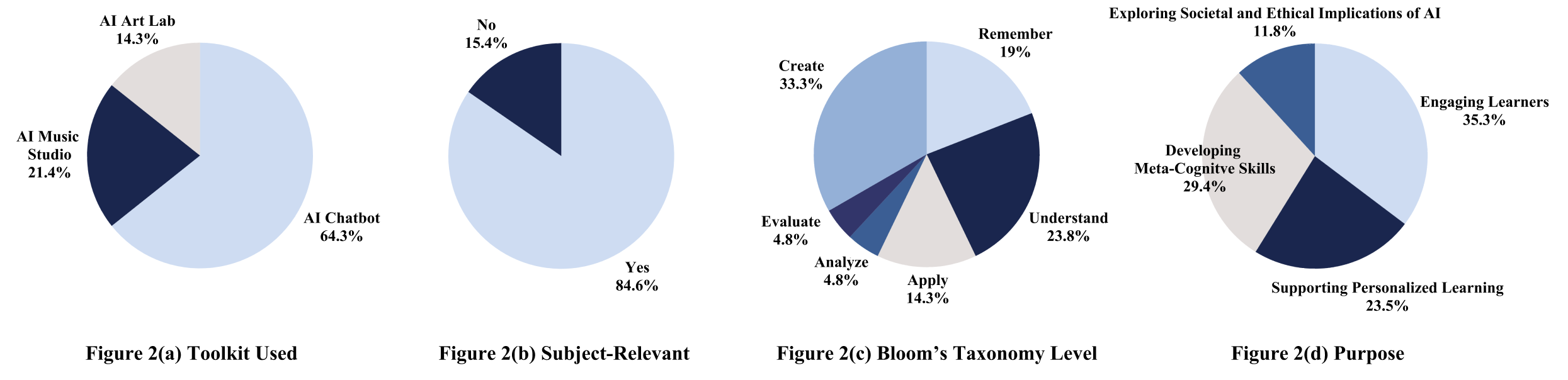} 
  \caption{Elements Included in Course Designs}
  \label{fig:distribution}
\end{figure*}

Thematic coding has been performed to understand the types of courses designed by teachers based on toolkit used Figure (\ref{fig:distribution}(a)), whether the course is relevant to what the teacher teaches Figure (\ref{fig:distribution}(b)), Bloom's Taxonomy level applied Figure (\ref{fig:distribution}(c)), and the purpose of the course Figure (\ref{fig:distribution}(d)). Regarding the \textbf{types of toolkit} used, AI chatbot has the widest adoption (n=9). For \textbf{interdisciplinary course-integration}, except for a single interdisciplinary teacher (V-MA-Z) who designed a course irrelevant to the course they teach, all other designs demonstrated how the toolkit can be integrated into traditional subjects. Aligning to \textbf{Bloom's Taxonomy} \cite{forehand2010bloom}, many teachers created activity to ask students create songs, stories, posters (create-level, 33.3\%); help students interact with AI for concept understanding (23.8\%); memorize concepts through AI-generated songs (remember-level, 19\%). The skills analysis and evaluation are relatively less involved in course designed. Lastly, the motivation of designing AI-enhanced activities including engaging learners (e.g. memorizing concepts in a more fun way, 35.3\%), supporting personalized learning (e.g. tutoring by chatbot, 23.5\%), developing meta-cognitive skills (e.g. problem-solving, critical thinking, 29.4\%), and exploring societal and ethical implications of AI (e.g. facilitating ethical discussions, 11.8\%). The diversity in most dimensions indicate the adaptivity of the toolkit.

\subsubsection{2.2 What are the desired and undesired features of the toolkit?} In this part, the most positive feedback we received focused on the \textbf{AI Prompt Evaluation Guidance}, its potential to assist learning, and the creative possibilities offered by the variety of tools and their outputs. V-MA-Z, who teaches high school media arts, believes that contemporary art, in particular, is a subject that requires strong supporting reasons and logical thinking. According to V-MA-Z, students could benefit from the prompt evaluation guidance to strengthen their logical reasoning skills.

\begin{quote}
    \scriptsize\textit{V-MA-Z: “I like how it [prompt evaluation] can provide suggestions for the students' questions, helping them to express their questions more precisely. It's something I think that benefits their mindset.”}
\end{quote}

P-GSci-Y sees the potential of the toolkit in \textbf{supporting student self-tutoring and reinforcing their knowledge}. For example, P-GSci-Y could set the AI chatbot to focus on geometry, allowing students to learn about different shapes by asking questions and receiving immediate answers. Additionally, other AI models, such as the Music Studio, can help students review interdisciplinary knowledge through the creative process. P-GS-Z also expressed appreciation for the prompt evaluation feature, stating that it promotes metacognitive while learning and enhances students' problem-solving abilities. 
Another reason students could benefit from \textbf{self-assisted learning} with AI is that AI serves as a 24/7 tutor. For example, P-GSci-X, whose students are primarily from low-income families, emphasized how the AI Chatbot could be a valuable tool for students who don't have access to someone knowledgeable in physics or math at home.

\begin{quote}
    \scriptsize\textit{P-GSci-X: “Most of my students don't have someone in the family who's like, 'Let me look at your math homework. Let me look at your physics homework.' Most of my students, it's like that's already beyond the people that they have access to at home.”}
\end{quote}

Additionally, the variety of AI tools was well-received by teachers. C-MathI-Z noted that the AI Art Lab and AI Music Studio expose students to more possibilities of AI beyond just chatbots. C-MathI-Z also praised the examples provided in the toolkit, particularly how the rubric is set up and filled out during the process, and found the technical level of the tools to be well-suited for high school students.

\begin{quote}
    \scriptsize\textit{C-MathI-Z : “I’m a big fan of the fact that it’s not just like the chat bot... and not shown the other things that AI does, which is, you know, generate art, or generate lyrics or music.”}
\end{quote}

Other teachers highlighted the \textbf{customization and rubric integration features}, such as scenario customization and the ability to integrate and break down rubrics. Additionally, the toolkit's ability to encourage student self-driven exploration was highly desired by the teachers.

While teachers appreciated certain features, they also expressed concerns about AI-generated content. Their main worries include \textbf{accuracy, trustworthiness, and the potential for irrelevant or unreliable content.} Copyright issues and content legitimacy were also mentioned. P-GSci-Y, with experience using AI models like ChatGPT, emphasized that AI-generated content can lack credibility, raising concerns about providing students with inaccurate information.
Other concerns towards the AI generated content appears to be more on legal issues, such as the copyright and compliance with international law. 

15\% of the teachers found that some tools in the toolkit did not align with their teaching scope or class content. P-CH-Z, a Chinese teacher at two public schools in the district, remarked, “I don’t see much reason to use the AI Art Lab for language learning." Additionally, P-GS-Z mentioned that, for general classroom use, the toolkit lacks enough content or activities to make it a go-to resource, especially when the goal is to encourage student self-exploration.

Concerns about the \textbf{variation in students' knowledge level, AI proficiency, and instructors' AI literacy and technological skills} are also significant. P-CH-Z emphasized the need for adaptable modules to match students' varying Chinese proficiency levels.

\begin{quote}
    \scriptsize\textit{P-CH-Z: “[AI Art lab] For beginner learners, I would give a simple sentence to describe the picture, and for intermediate learners, a slightly more complex sentence. This helps learners choose based on their skill level.”} 
\end{quote}

C-MathI-Z suggested adding challenges for students already familiar with AI tools to keep them engaged and encourage deeper exploration. They also stressed the need for better onboarding support, including scaffolding for teachers, helpful prompts for students to create content, and comprehension questions to guide learning.

\begin{quote}
    \scriptsize\textit{P-GSci-Y:” For example, in the AI chatbot and AI art lab, it could assist teachers by suggesting topics or prompts, or by providing some templates.”}
\end{quote}

Other suggestions were more related to UX design, such as overwhelming text and excessive blank fields to fill in. Teachers expressed concerns that this kind of design might cause students to lose interest quickly. 

\subsubsection{2.3: How can teachers use such features to overcome difficulties and design courses?}

From teachers' answers to questions under \textit{2) AI Literacy teaching experiences}, difficulties teachers encountered in teaching could be summarized as the following 4 main categories: 

\begin{itemize}
    \item 1) Limited teaching resources
    \item 2) Students AI abilities gap
    \item 3) Lack of hands-on experience for students
    \item 4) Teacher AI literacy gap
\end{itemize}

When talking about whether our toolkit reduced such barriers, over 46\% of teachers mentioned that \textbf{using the PBL Toolkit could help solve one or more challenges} in the course plans they designed. For example, P-GSci-X's main difficulty is the limited teaching resources. As a general science teacher, P-GSci-X is responsible for teaching multiple subjects such as physics, medical science, engineering, and math. P-GSci-X explained that it is impossible for a teacher to monitor every student's work and answer all the questions that may arise. In the course plan designed by P-GSci-X, the AI chatbot serves as a conversational tutor, providing students with a platform to discuss physics concepts outside of class. This allows students to \textbf{practice in a non-judgmental, supportive environment}. Students not only reinforce their knowledge through conversations with AI but also solve questions that do not require the teacher's direct involvement.

\begin{quote}
    \scriptsize\textit{P-GSci-X: “ [note uploading process]...I would say, Hey, go ahead and open up the Chatbot, upload these notes and then ask it some questions. You know, to clarify any misunderstandings that you have before. You know before you talk to me about it. I'd say, you know. Spend 10 to 15 minutes reading the notes and reviewing them with an AI assistant.” }
\end{quote}

P-GSci-X suggested using AI to review student work, saving teachers time and providing valuable feedback. By taking a picture of their work, students could receive AI-generated insights, which is especially helpful for those without knowledgeable support at home.

A direct example of the \textbf{gap in students' AI abilities} can be seen in P-IT-Z’s class. P-IT-Z teaches IT at a school near big tech companies, with students from both tech and non-tech backgrounds. This creates a gap in AI proficiency: Some students complete tasks quickly, while others struggle with basic functions like logging in. P-IT-Z's course uses the AI Art Lab tool in design projects, where students create visuals for products they want to sell in an app. This approach encourages hands-on, self-directed learning. The AI Art Lab's task breakdown feature helps less proficient students, while advanced students can work at their own pace.

Teachers not only face challenges with varying levels of student AI literacy, but also encounter \textbf{gaps in their own technical abilities}. V-IT-Z, an IT teacher without a technical background, teaches middle and high school students. When using AI tools such as Midjourney, students with limited English and technical skills heavily rely on V-IT-Z, who struggles to explain complex concepts. To address this, V-IT-Z integrated an AI chatbot into the course, allowing students to explore independently. Teachers assign tasks like designing board games or orientation plans, and younger students use the chatbot to simplify complex terms, while older students use it for homework help, reducing their reliance on the teacher.

Although a PBL AI Toolkit can address many of the major difficulties teachers face in the classroom, \textbf{there are still some practical issues that fall outside its scope. } One of the challenges is the lack of widespread access to AI resources. Some teachers mentioned that classroom equipment and layout can affect how they teach AI.

\begin{quote}
    \scriptsize\textit{V-Bio-Z: “The biggest challenge is the hardware. Graphics cards are expensive, and the school's equipment just not enough... For a formal biology class, we’d need every student to have a tablet to use the tools, or we’d have to move the whole class to the computer lab, which is a bit of a hassle."}
\end{quote}

\subsection{RQ3: How do differences in teacher and student backgrounds influence their perspectives on AI tools, student performance, and course design?}
We compare similarities, differences, and patterns of variation between student and instructor backgrounds to identify similarities and differences between demographics. We present both horizontal and vertical comparisons to explore how these factors relate to potential AI Literary courses and tools design and future development.

\subsubsection{Comparison 1: Between student group from different family income levels.} Among the interviewees, P-GSci-X and P-GS-XY teach at schools where most students are from low-income families, with P-GS-XY also teaching students from diverse backgrounds like gig worker households and single-parent families. One teacher works with new immigrants but did not mention low-income situations. The remaining 10 teachers primarily teach students from middle- and upper-middle-class families. We categorized the teachers into three groups: Group A (low-income students), Group B (middle- and upper-middle-class students), and Group C (mixed backgrounds). \\
\begin{quote}
    \textit{\textbf{Group A}}: P-GSci-X and P-GS-XY \\ 
    \textbf{\textit{Group B}}: P-GS-Z, V-MA-Z, V-IT-Z, V-IT-Z-2\\
    \textbf{\textit{Group C}}: C-MathI-Z, P-CH-Z, P-SS-Z, P-GSci-Y, P-IT-Z, V-Bio-Z, V-EngIT-Z\\
\end{quote}

\subsubsection{Insight 1: Potential AI Literacy Equity through Teachers' AI Literacy Level and Instruction Offered. }Although there is a difference in students' AI usage between groups (nearly 100\% of Group B's students vs. 50\% of Group A's students used AI tools), within the range of this study, we did not find any connection between students' economical background and teachers' AI Literacy level, teaching experiences, and attitudes towards AI Literacy. Not only do teachers' self-rated AI literacy level not correlated with their students' economical status, all teachers in Group A and Group B, and 4 out of 7 teachers in Group C had AI Literacy teaching experiences in AI-integrated traditional subject, summer camps, AI-focused courses, and so on. In other words, such messages encourage us to hold an optimistic attitude towards AI Literacy exposure to learners with varied economical background. 

Although teachers generally have positive views on the use of AI for learning, highlighting its growing relevance and support for self-directed learning, they also raised several concerns that can possibly be affected by students' or schools' economic status. A common theme across all three groups was the challenge of bridging the gap in AI proficiency and technical skills among students. For instance, P-GSci-X mentioned that \textit{“many students don't have access to the Internet at home", “90\% of students interacted with Gemini, as it shows as Google results; the other AI applications require registration that our students may not know how to do".} In addition, teachers emphasized the need for proper AI training to enhance their own understanding. Groups A and B specifically identified a technological literacy gap among educators as a significant issue. Lastly, limited resources, both in terms of instructional time and access to devices within and outside of the school, further restrict efforts to advance AI literacy education.

\subsubsection{Comparison 2: Between Instructor’s Different Literacy Levels} 
In this section, we divided our interviewees based on their self-rated scores from part 1: “How confident are you in understanding AI results and knowing their limits  (out of 100\%) ? " The median score from this question was approximately 58.46\%. We split the participants into two groups: teachers with a confidence level higher than 58.46\% were categorized as having high AI literacy and assigned to Group X, while those with a confidence level lower than 58.46\% were categorized as having low AI literacy and assigned to Group Y.
The grouping is as follows:\\

\begin{quote}
    \textit{\textbf{Group X}}: P-GS-Z, V-MA-Z, P-GSci-X, P-SS-Z, P-GS-XY, V-EngIT-Z\\
    \textbf{\textit{Group Y}}: C-MathI-Z, P-CH-Z, P-GSci-Y, V-IT-Z, P-IT-Z, V-IT-Z-2 , V-Bio-Z\\
\end{quote}



\textbf{Insight 2: Teachers' activity design skills and topics are not specifically related to either their self-reported AI Literacy level or teaching experience.} Group X, with higher AI proficiency and an average of 8 years experience, focuses on advanced skills like critical thinking and AI's capabilities, using tools like AI Chatbot and AI Art Lab to overcome teaching challenges. Group Y, with slightly less experience (7.7 years) and lower AI literacy, emphasizes ethical AI use and applies tools like Chatbot and music-based AI for memory retention, though with mixed success. Overall, higher proficiency teachers prioritize critical thinking, while lower proficiency teachers focus on ethical AI use and literacy. Both groups aim to help students use AI effectively while developing soft skills such as problem-solving. 

%% file: Sections/06_Discussion.tex
\section{Discussion}
Our results highlight several key points about integrating the PBL AI Literacy toolkit into K-12 education. First, while many teachers are confident in understanding AI's basic capabilities, \textbf{they still face challenges, especially with students' varying AI skill levels.} Teachers with higher AI literacy were better able to integrate the toolkit into their lesson plans, using tools like the AI Chatbot to foster independent learning and critical thinking. However, those with lower AI literacy tended to focus more on ethical concerns and proper AI use, both of them lack of AI educational knowledge, such as knowing the AI for K-12 framework and 5 big ideas. indicating a need for more training and support  for both students and instructors in AI education.
Another takeaway is that \textbf{students' opportunities to learning AI Literacy at school might not be significantly varied by their economic status.} Instructors teaching lower-income students showed success in guiding their students properly through AI tools and produce satisfying study results, despite resource constraints, which challenges assumptions about access and technological inequality.

Although the toolkit offers adaptivity, flexibility, and creativity, concerns about the dependency of the AI tool, the accuracy of AI-generated content, and legal issues such as copyrights need to be addressed. Teachers also raised the need for more support and onboarding, particularly in using more advanced features.

There are limitations in this research that should be considered when interpreting the findings, including a small sample size and limited diversity, which may affect the generalizability of the findings. Future studies should involve a broader demographic for a more comprehensive understanding of AI literacy challenges. Future work could also focus on improving teacher training to enhance AI literacy, offering clearer classroom guidance, addressing ethical concerns, and making the toolkit scalable for diverse educational settings.